\documentclass[wcp]{jmlr}

\usepackage{longtable}

\usepackage{booktabs}

\makeatletter
\let\Ginclude@graphics\@org@Ginclude@graphics 
\makeatother

\jmlrvolume{222}
\jmlryear{2023}
\jmlrworkshop{ACML 2023}

\usepackage{xcolor}

\title[BarlowRL: Barlow Twins for Data Efficient Reinforcement Learning]{BarlowRL: Barlow Twins for Data-Efficient Reinforcement Learning}

\author{\Name{Omer Veysel Cagatan} \Email{ocagatan19@ku.edu.tr} \\
\Name{Baris Akgun} \Email{baakgun@ku.edu.tr} \\
\addr {Koc University, Turkey}
}

\editors{Berrin Yan{\i}ko\u{g}lu and Wray Buntine}

\begin{document}

\maketitle

\begin{abstract}
This paper introduces BarlowRL, a data-efficient reinforcement learning agent that combines the Barlow Twins self-supervised learning framework with DER (Data-Efficient Rainbow) algorithm. BarlowRL outperforms both DER and its contrastive counterpart CURL on the Atari 100k benchmark. BarlowRL avoids dimensional collapse by enforcing information spread to the whole space. This helps RL algorithms to utilize uniformly spread state representation that eventually results in a remarkable performance. The integration of Barlow Twins with DER enhances data efficiency and achieves superior performance in the RL tasks. BarlowRL demonstrates the potential of incorporating self-supervised learning techniques, especially that of non-contrastive objectives, to improve RL algorithms.

\end{abstract}

\section{Introduction}

Deep reinforcement learning (RL) has made significant strides in achieving impressive accomplishments, including surpassing human or super-human performance in various domains such as complex games, for example as demonstrated by OpenAI Five~\cite{Berner2019Dota2W}, AlphaGo ~\cite{Silver2016MasteringTG}, and AlphaStar ~\cite{Vinyals2019GrandmasterLI}. However, these achievements have relied on the utilization of large-scale neural networks and an immense number of environment interactions for training. For example, to accumulate equivalent experience as OpenAI Five or AlphaGo, a human player would require tens of thousands of years of gameplay. The use of large networks is often necessary to capture the complexity of the problem and allow expressive value estimation and policy representation. Moreover, a substantial number of samples is crucial for effectively training such networks and determining the long-term effects of different action choices. Despite these successes, achieving human-level sample efficiency in RL remains an ongoing and prominent objective in the field.

Along this line, we focus on the Atari 100k benchmark ~\cite{Kaiser2019ModelBasedRL} which is a widely used evaluation framework for assessing the performance of RL algorithms on a diverse set of Atari 2600 games. It consists of a collection of 100,000 human-normalized game episodes, each represented as a sequence of observations, actions, and rewards. The benchmark serves as a standard testbed to measure the effectiveness of RL algorithms in learning policies that achieve high scores in these games. The complexity of the Atari 100k benchmark arises from the diverse game dynamics, varying reward structures, and the need for agents to exhibit robust generalization across different games. RL algorithms are evaluated based on their ability to learn efficient policies, improvement over time, and 
performance compared to human players. The benchmark provides a comprehensive and challenging environment to evaluate and compare the capabilities of different RL approaches in the domain of Atari games. 

~\cite{Schwarzer2020DataEfficientRL} demonstrate that representation learning is a major step towards getting closer to the human-level performance in Atari 100k by predicting its own latent state representations multiple steps into the future. To promote the exploration of non-contrastive objectives in representation learning for RL, we introduce BarlowRL, a framework that minimizes the overhead in terms of architecture and model learning. BarlowRL utilizes the Barlow Twins objective~\cite{Zbontar2021BarlowTS}, operating within the same latent space and architecture commonly employed in model-free RL. This integration seamlessly fits into the training pipeline, eliminating the need for additional hyperparameters. BarlowRL outperforms its baseline DER~\cite{Hasselt2019WhenTU} and its contrastive counterpart CURL~\cite{Srinivas2020CURLCU} by a significant margin in all metrics. 

Our paper highlights the key contributions of BarlowRL, presenting it as a straightforward performant framework that effortlessly combines non-contrastive learning with model-free RL. Importantly, this integration requires minimal alterations to the existing architectures and training pipelines.

\section{Related Work}

Sample efficiency has always been a crucial aspect of RL evaluation due to the high cost of interacting with an environment.~\cite{Kaiser2019ModelBasedRL} introduced the Atari100K benchmark, which has proven to be valuable for assessing sample efficiency and has spurred recent advancements.~\cite{Kostrikov2020ImageAI} utilized data augmentation to develop a sample-efficient RL method called DrQ, surpassing previous approaches on Atari100K. \cite{Srinivas2020CURLCU} add a contrastive auxiliary loss atop a off-policy reinforcement learning. Data-Efficient Rainbow (DER)~\cite{Hasselt2019WhenTU} and DrQ($\epsilon$) ~\cite{Agarwal2021DeepRL} achieved superior performance by simply adjusting hyperparameters of existing model-free algorithms, without introducing any algorithmic innovations. ~\cite{Schwarzer2020DataEfficientRL} introduced SPR which incorporated a self-supervised temporal consistency loss based on BYOL~\cite{Grill2020BootstrapYO} along with data augmentation. SR-SPR ~\cite{Nikishin2022ThePB} combined SPR with periodic network resets to achieve state-of-the-art performance on the Atari100K benchmark. EfficientZero ~\cite{Ye2021MasteringAG}, an efficient variant of MuZero ~\cite{Schrittwieser2019MasteringAG}, learned a discrete-action latent dynamics model from environment interactions and employed lookahead MCTS in the latent space.~\cite{Micheli2022TransformersAS} introduced IRIS, a data-efficient agent that learns in a world model composed of an autoencoder and an autoregressive Transformer. BBF ~\cite{Schwarzer2023BiggerBF} is an RL agent that attains super-human performance on the Atari100K benchmark. The key strategy employed by BBF involves scaling the neural networks utilized for value estimation. In addition to network scaling, BBF incorporates various design choices that facilitate efficient learning from limited samples, leading to improved performance. 

\begin{figure}[htbp]
  \centering
  \includegraphics[width=0.60\textwidth]{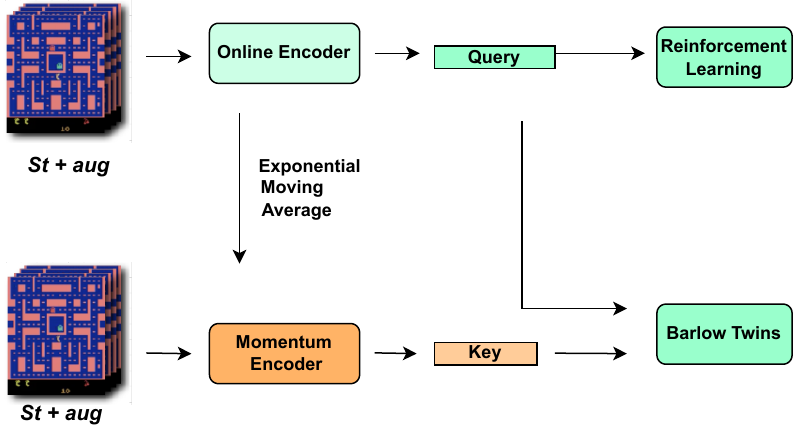}
  \caption{BarlowRL is a framework that combines non-contrastive learning and reinforcement learning. It trains a visual representation encoder by ensuring that the embeddings of the data-augmented versions of anchor observations (query) and key observations (key) match using a non-contrastive loss. These examples are constructed from the minibatch sampled for the reinforcement learning update. The keys are encoded with a momentum-averaged version of the query encoder. The reinforcement learning policy and/or value function take query encoder output as input, which is jointly trained with the non-contrastive and reinforcement learning objectives.}
  \label{fig:arc}
\end{figure}

\section{Background}
\subsection{Rainbow}

RainbowDQN ~\cite{Hessel2017RainbowCI} can be understood as an amalgamation of various enhancements built upon the original DQN ~\citep{Mnih2015HumanlevelCT}. The DQN approach utilize Q-Learning, an off-policy algorithm, and a convolutional neural network to approximate action value functions by mapping raw pixels. Subsequently, several improvements have been proposed, including Double Q-Learning ~\citep{Hasselt2015DeepRL}, Dueling Network Architectures ~\citep{Wang2015DuelingNA}, Prioritized Experience Replay~\citep{Schaul2015PrioritizedER}, Noisy Networks~\citep{Fortunato2017NoisyNF} and distributional reinforcement learning~\citep{Bellemare2017ADP} with the C51 Algorithm, which predicts a distribution over potential value function bins.

RainbowDQN integrates all these techniques into a unified off-policy algorithm, exhibiting state-of-the-art sample efficiency on Atari benchmarks. Furthermore, RainbowDQN incorporates the utilization of multi-step returns ~\cite{Sutton1998IntroductionTR}. A data-efficient variant of RainbowDQN was proposed by ~\cite{Hasselt2019WhenTU}, which entails an improved configuration of hyperparameters optimized for performance benchmarked at 100K interaction steps.

\subsection{Contrastive Learning}
Contrastive Learning serves as a framework for acquiring representations by incorporating similarity constraints within a dataset, typically organized into pairs of similar and dissimilar instances. This framework can be likened to a dictionary lookup task, where positive and negative pairs represent sets of keys in relation to a query or anchor. An example of contrastive learning is Instance Discrimination~\cite{Wu2018UnsupervisedFL}, where positive pairs consist of data augmentations of the same instance (e.g., image), while negative pairs involve different instances. The selection of negative pairs poses a significant challenge in contrastive learning, as it strongly influences the quality of the learned representations. Different loss functions, such as InfoNCE~\cite{Oord2018RepresentationLW}, Triplet~\cite{Wang2015UnsupervisedLO}, Siamese~\cite{Chopra2005LearningAS} among others, can be employed for contrasting purposes.

\subsection{Non-contrastive Learning}
Recent advancements in the field of visual self-supervised learning (visual SSL) have moved beyond the traditional contrastive paradigm and have found ways to reduce the reliance on negative samples. These new approaches focus on optimizing the affinity of augmented representations alone and fall under the category of non-contrastive frameworks. To prevent model collapsing, several common practices have emerged, including the use of asymmetrical architecture ~\cite{Grill2020BootstrapYO,Chen2020ExploringSS}, dimension de-correlation~\cite{Zbontar2021BarlowTS,Bardes2021VICRegVR,Ozsoy2022SelfSupervisedLW,Ermolov2020WhiteningFS}, and clustering~\cite{Amrani2021SelfSupervisedCN,Assran2022MaskedSN,Caron2018DeepCF,Caron2021EmergingPI}.

\subsection{Barlow Twins}
Barlow Twins~\citep{Zbontar2021BarlowTS} is a self-supervised learning method that aims to learn meaningful representations from unlabeled data without the need for manual annotation or labels.

The key idea behind Barlow Twins is to maximize the agreement between two augmented views of the same input while simultaneously minimizing the agreement across different inputs. This is achieved by introducing a redundancy reduction objective that encourages the representations of similar inputs to become more similar and the representations of dissimilar inputs to become more dissimilar.

Barlow Twin loss is defined as follows: 

\begin{equation}
\mathcal{L_{BT}} \triangleq \sum_i (1-\mathcal{C}_{ii})^2 + \lambda \sum_{i}\sum_{j \neq i}{\mathcal{C}_{ij}}^2 
\end{equation}

where \(\lambda > 0\)  balances invariance (diagonal) and redundancy reduction (off-diagonal) terms. \(\mathcal{C}\) is the cross-correlation matrix from identical network embeddings and the subscripts index the rows and columns. The individual elements of this matrix, \(\mathcal{C}_{ij}\), range between -1 (anti-correlation) and 1 (perfect correlation) and are calculated as follows:
 
\begin{equation}
\mathcal{C}_{ij} \triangleq \frac{\sum_b z^A_{b,i} z^B_{b,j}}{\sqrt{\sum_b {(z^A_{b,i})}^2} \sqrt{\sum_b {(z^B_{b,j})}^2}}
\label{eq:covar}
\end{equation}

In Eq.~\ref{eq:covar}, \(b\) represents an individual sample in the current batch, \(i\) and \(j\) are dimension indices of the embeddings' output. Furthermore, \(z^A_{b,i}\) and \(z^B_{b,j}\) are the elements of embedding vectors from two identical networks for the \(i^{th}\) and \(j^{th}\) dimensions respectively, corresponding to the \(b^{th}\) sample in the batch.
\subsection{CURL}
The CURL (Contrastive Unsupervised Representations for Reinforcement Learning)~\cite{Srinivas2020CURLCU} approach is a deep reinforcement learning framework that combines unsupervised representation learning with reinforcement learning. It aims to learn useful representations from raw sensory inputs, such as images or pixels, in a self-supervised manner, without requiring explicit labels or rewards.

At its core, CURL  leverages the principles of contrastive learning to learn representations. Contrastive learning is a self-supervised learning technique where the model learns to differentiate between positive pairs (similar samples) and negative pairs (dissimilar samples) that comes from the batch. By maximizing the similarity between positive pairs and minimizing the similarity between negative pairs, the model can learn meaningful representations that capture important features and patterns in the data.

In the context of reinforcement learning, CURL extends contrastive learning to enable effective exploration and generalization in complex environments. It combines a contrastive objective with a reinforcement learning objective to jointly optimize the representation learning and policy learning processes.

\section{Method}
BarlowRL is a modified version of a base reinforcement learning (RL) algorithm that incorporates a non-contrastive objective as an auxiliary loss during batch updates. In our experiments, we trained BarlowRL alongside a model-free RL algorithm: Rainbow DQN (data-efficient version) for Atari experiments.

\subsection{Architecture}
BarlowRL uses Barlow Twins objective as introduced by \cite{Zbontar2021BarlowTS}. In most Deep RL architectures, a common approach is to use a sequence of consecutive frames as input~\citep{Hessel2017RainbowCI}. As a result, the non-contrastive objective is carried out across these frame stacks instead of individual image instances. Even though momentum encoding was not utilized by ~\cite{Zbontar2021BarlowTS}, we keep it as done by~\cite{Srinivas2020CURLCU}. By using Polyak averaging, the target network becomes a slower changing version of the online network. This stability promotes a more consistent estimation of the Q-values and helps to prevent overestimation or underestimation biases that can occur during training~\citep{Mnih2015HumanlevelCT}. The non-contrastive representation is trained in conjunction with the RL algorithm, and the latent code receives gradients from both the non-contrastive objective and the Q-function. An illustration of our architecture is presented in Figure \ref{fig:arc}.

\subsection{Positive Pair Generation}
As in the non-contrastive computer vision approaches~\citep{Zbontar2021BarlowTS,Bardes2021VICRegVR,Ozsoy2022SelfSupervisedLW,Grill2020BootstrapYO}, we only need positive samples, unlike the contrastive objectives which require batch negatives~\citep{Srinivas2020CURLCU}. BarlowRL only utilizes random cropping as in ~\cite{Srinivas2020CURLCU} since the comparison of non-contrastive and contrastive objectives is our primary goal.

A notable distinction between RL and computer vision settings is that in model-free RL algorithms operating from pixels, an input instance is a stack of frames~\citep{Mnih2015HumanlevelCT}, instead of a single image.
\section{Experiments}

In this work, we focus on the Atari 100k benchmark, which is a well-established benchmark that has been used to evaluate a variety of sample-efficient reinforcement learning algorithms. In Section \ref{section:base}, we present the baselines. In Section \ref{section:res}, we describe the evaluation protocol and discuss the results of our experiments.

\subsection{Baselines}
\label{section:base}

Atari 100k is widely regarded as one of the most popular benchmarks for evaluating RL agents. Each agent incorporates its own set of premises and introduces different novelties. Due to the various design choices, it is important to separate and compare them based on their specific characteristics. In our evaluation, we categorize the agents into five distinct categories:
lookahead search assisted agents ~\citep{Ye2021MasteringAG,Schrittwieser2019MasteringAG}, pre-trained agents~\citep{Schwarzer2021PretrainingRF}, model-based agents~\citep{Kaiser2019ModelBasedRL,Micheli2022TransformersAS}, modified model-free agents~\citep{Schwarzer2023BiggerBF,Agarwal2021DeepRL,Nikishin2022ThePB}, and baseline model-free agents~\citep{Hasselt2019WhenTU,Schwarzer2020DataEfficientRL,Srinivas2020CURLCU}.

To further elaborate on our distinctions, we describe the work by ~\cite{Nikishin2022ThePB} as a modified model free agent because it adds periodic resets to some layers of the agent, which uses the approach of ~\cite{Schwarzer2020DataEfficientRL} as its baseline. ~\cite{Schwarzer2023BiggerBF} enhances SPR by scaling it to a larger number of parameters. 

Among the model-free approaches, SPR~\citep{Schwarzer2020DataEfficientRL} introduces a notable distinction by employing bootstrapping techniques on latent representations derived from future states. In contrast, other models, including BarlowRL, rely exclusively on augmented views of the present state for representation learning. SPR generates multiple predictions for future states, which are subsequently processed through a transition model and projection head to apply self-supervised consistency loss or BYOL similarity. The prediction of future states from the present is inherently significant in reinforcement learning, as it informs an agent's action selection based on current state information. 

In our evaluation, we specifically focus on model-free agents as they play a crucial role in the field of reinforcement learning. We use DER~\citep{Hasselt2019WhenTU}, CURL~\citep{Srinivas2020CURLCU}, and DrQ~\citep{Kostrikov2020ImageAI} as our main baselines. DER and DrQ comparison would highlight the benefits of adding representation learning and CuRL comparison would highlight the advantages of a noon-contrastive loss over a contrastive loss. We also include SPR which additionally utilizes future states for representation learning, instead of just the current state like BarlowRL and others. As such, do not take it as a direct competitor but present its results to show how BarlowRL performs with such a disadvantage.

While there are various types of agents that have demonstrated impressive performance on the Atari 100k benchmark, we deliberately narrow down our analysis to these model-free approaches to effectively see the impact of representation learning, and compare contrastive and non-contrastive objectives. The reason behind this decision is that our primary objective is to introduce BarlowRL as a novel baseline agent, rather than directly comparing it with state-of-the-art methods. By exclusively considering model-free agents, we aim to provide a comprehensive assessment of BarlowRL's performance within the context of existing approaches that rely on model-free reinforcement learning techniques or agents enhanced by representation learning methods. 

\begin{figure}[htbp]
  \centering
  \includegraphics[width=1\textwidth]{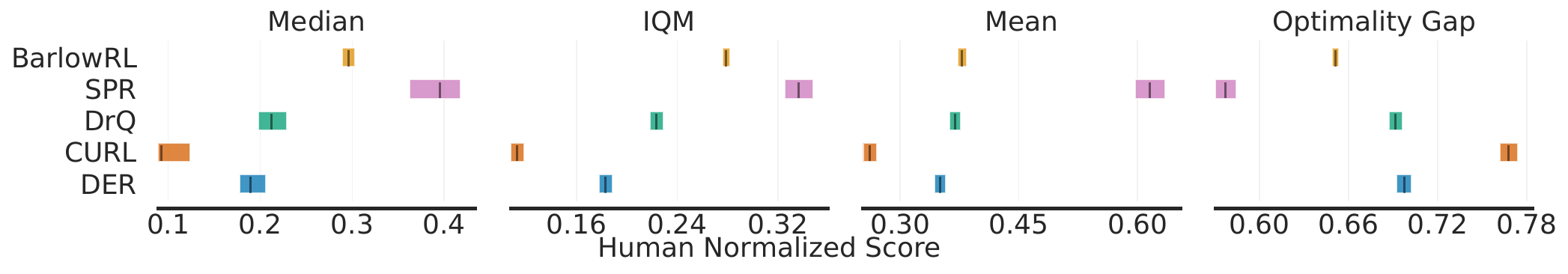}
  \caption{Mean, median, interquartile mean human normalized scores and optimality gaps (lower is better), computed with stratified bootstrap confidence intervals for 50 runs for BarlowRL, and 100 runs for SPR, CURL, and DrQ ~\citep{Agarwal2021DeepRL}.}
  \label{fig:metric}
\end{figure}

\begin{table}[t]
    \caption{Returns on the 26 games of Atari 100k after 2 hours of real-time experience, and human-normalized aggregate metrics. SPR is excluded from the comparison as discussed in Section~\ref{section:base}.}
    \label{tab:atari_results_full}
\begin{center}
\begin{small}
\centering
\scalebox{0.82}{
\centering
\begin{tabular}{lrrrrrr|l}
\toprule
Game                 &  Random    &  Human    &  DER     &  DrQ    & CURL     &\textsc{BarlowRL} (ours) &  SPR        \\
\midrule
Alien                &  227.8     &  7127.7   &  \textbf{802.3}   &  734.0  & 711.0    &  734.9                  &  841.9       \\
Amidar               &  5.8       &  1719.5   &  125.9   &  94.2   & 113.7    &  \textbf{189.9}                  &  179.7       \\
Assault              &  222.4     &  742.0    &  561.5   &  479.5  & 500.9    &  \textbf{796.2}                  &  565.6           \\
Asterix              &  210.0     &  8503.3   &  534.5   &  535.6  & 567.2    &  \textbf{848.2}                  &  962.5     \\
BankHeist            &  14.2      &  753.1    &  185.5   &  153.4  & 65.3     &  \textbf{466.5}                 &  345.4      \\
BattleZone           &  2360.0    &  37187.5  &  8977.0  &  10563.6& 8997.8   &  \textbf{19322.0}                &  14834.1     \\
Boxing               &  0.1       &  12.1     &  -0.3    &  \textbf{6.6}    & 0.9      &  0.3                    & 35.7    \\
Breakout             &  1.7       &  30.5     &  9.2     &  \textbf{15.4}   & 2.6      &  2.6                    & 19.6       \\
ChopperCommand       &  811.0     &  7387.8   &  925.9   &  792.3  & 783.5    &  \textbf{1186}                   & 946.3        \\
CrazyClimber         &  10780.5   &  35829.4  &  \textbf{34508.6} &  21991.5& 9154.4   &  23250.8                & 36700.5       \\
DemonAttack          &  152.1     &  1971.0   &  627.6   &  \textbf{1142.4} & 646.5    &  868.9                  & 517.6        \\
Freeway              &  0.0       &  29.6     &  20.9    &  17.7   & 28.3     &  \textbf{27.6}                   & 19.3      \\
Frostbite            &  65.2      &  4334.7   &  871     &  508.0  & 1226.5   &  \textbf{2350.4}                 & 1170.7       \\
Gopher               &  257.6     &  2412.5   &  467.0   &  \textbf{618.0}  & 400.9    &  308.1                  & 660.6   \\
Hero                 &  1027.0    &  30826.4  &  6226.0  &  3722.6 & 4987.7   &  \textbf{7163.1}                 & 5858.6    \\
Jamesbond            &  29.0      &  302.8    &  275.7   &  251.7  & 331.0    &  \textbf{345.2}                  & 366.5    \\
Kangaroo             &  52.0      &  3035.0   &  581.7   &  974.4  & 740.2    & \textbf{1431.6}                 & 3617.4        \\
Krull                &  1598.0    &  2665.5   &  3256.9  &  \textbf{4131.3} & 3049.2   &  3177.4                 & 3681.6            \\
KungFuMaster         &  258.5     &  22736.3  &  6580.1  &  7154.5 & 8155.6   &  \textbf{10716.4}                & 14783.2    \\
MsPacman             &  307.3     &  6951.6   &  1187.4  &  1002.9 & 1064.0   &  \textbf{2340.0}                 & 1318.4      \\
Pong                 &  -20.7     &  14.6     &  \textbf{-9.7}    &  -14.2  & -18.5    &  -16.9                  & -5.4               
 \\
PrivateEye           &  24.9      &  69571.3  &  72.8    &  24.8   & 81.9     &  \textbf{86.8}                   & 86             \\
Qbert                &  163.9     &  13455.0  &  \textbf{1773.5}  &  934.2  & 727.0    &  851                    & 866.3     \\
RoadRunner           &  11.5      &  7845.0   &  \textbf{11843.4} &  8724.6 & 5006.1   &  2954.6                 & 12213.1      \\
Seaquest             &  68.4      &  42054.7  &  304.6   &  310.4  & 315.2    &  \textbf{505.2}                  & 558.1       \\
UpNDown              &  533.4     &  11693.2  &  3075.0  &  3619.1 & 2646.4   &  \textbf{3725.9}                 & 10859.2       \\
\midrule
\#Superhuman (↑)     &  0         &  N/A       &  2      &  3      & 2        &  3         & 6          \\
Mean (↑)             &  0.000     &  1.000     &  0.350  &  0.369  & 0.261 &  \textbf{0.378}    & 0.616         \\
Median (↑)           &  0.000     &  1.000     &  0.189  &  0.211  & 0.092    &  \textbf{0.296}     & 0.396                   \\
IQM (↑)              &  0.000     &  1.000     &  0.183  &  0.223  & 0.113    &  \textbf{0.278}     & 0.337           \\
Optimality Gap (↓)   &  1.000     &  0.000     &  0.697  &  0.691   & 0.768    &  \textbf{0.651}     & 0.691           \\
\bottomrule
\end{tabular}
 }
\end{small}
\end{center}
\vspace{-0.02\linewidth}
\end{table}

\subsection{Results}
\label{section:res}

We use the human normalized score, which is commonly used metric in the Atari 100k benchmark, for comparing methods across multiple games. It is defined as follows:

\begin{equation}
hns = \frac{score\_{agent} - score\_{random}}{score\_{human} - score\_{random}}
\end{equation}
where \textit{score\_random} is the score obtained by a random policy, and \textit{score\_human} is obtained from human players by \cite{Wang2015DuelingNA}. 

In their work, ~\cite{Agarwal2021DeepRL} address the limitations of using mean and median scores as point estimates in RL benchmarks. They demonstrate that significant discrepancies can arise between these standard point estimates and interval estimates, which can impact the reliability and interpretability of benchmark results. 

\begin{figure}[tb]
  \centering
  \includegraphics[width=1\textwidth]{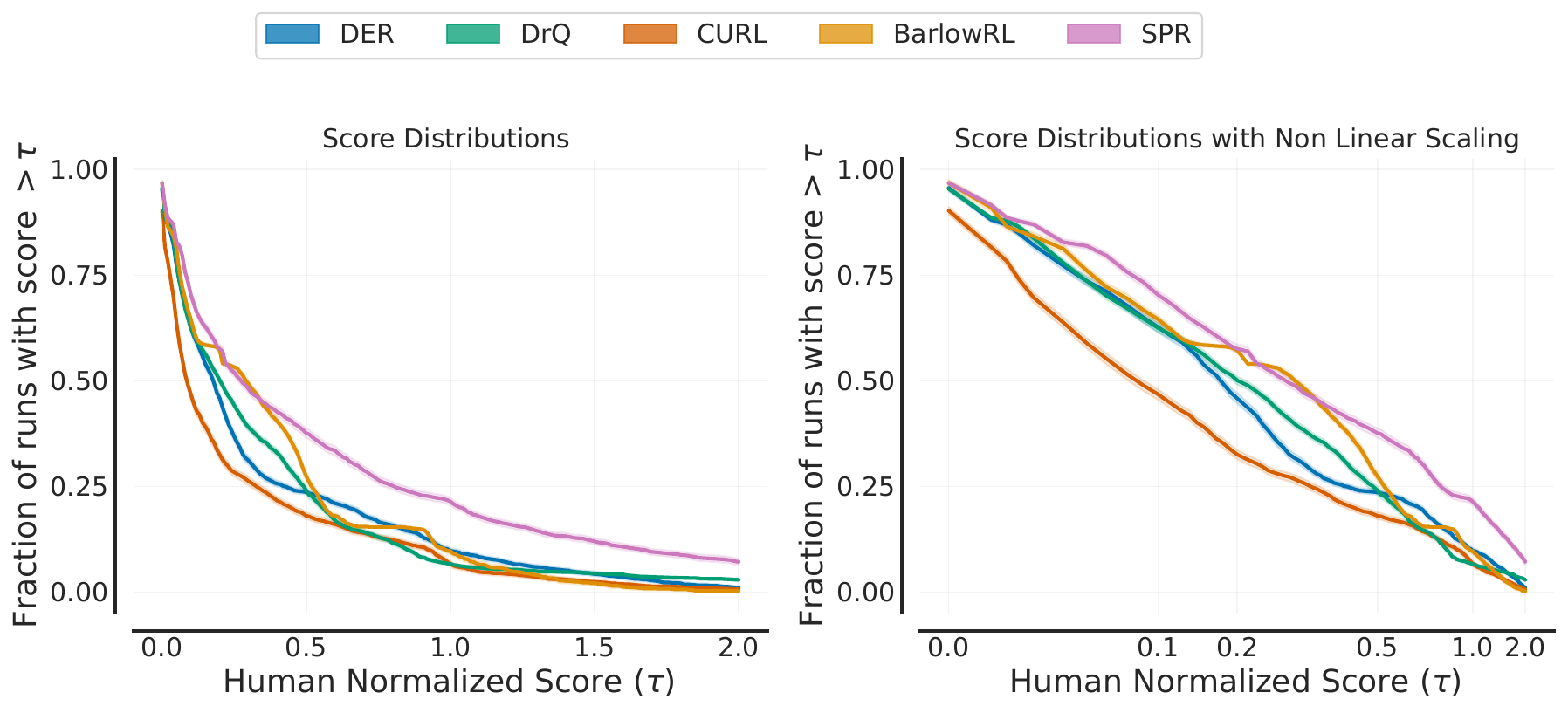}
  \caption{Performance profiles, i.e., fraction of runs above a given human normalized score.}
  \label{fig:score}
\end{figure}

Following the recommendations provided by ~\cite{Agarwal2021DeepRL}, we present human-normalized aggregate metrics along with returns across games in Table \ref{tab:atari_results_full}. In addition, we present a summary of human-normalized scores in Figure \ref{fig:metric}. To account for the uncertainties, we use stratified bootstrap confidence intervals for the mean, median, interquartile mean (IQM) and optimality gap. By incorporating these confidence intervals, we provide a more comprehensive and robust representation of the performance scores, allowing for more accurate comparisons. Furthermore, to facilitate finer comparisons, we also present performance profiles in Figure \ref{fig:score}. In all the table and the figures, we report the results published by~\cite{Agarwal2021DeepRL} for CURL, DrQ and SPR, and \cite{Schwarzer2023BiggerBF} for DER. Finally, we evaluate BarlowRL by computing an average of over 100 episodes collected at the end of training for each game (50 runs). 

In terms of performance across multiple metrics, BarlowRL has exhibited superiority over DER~\citep{Hasselt2019WhenTU}, CURL~\citep{Srinivas2020CURLCU}, and DrQ~\citep{Kostrikov2020ImageAI}, with the exception of SPR~\citep{Schwarzer2020DataEfficientRL}. BarlowRL's performance over DER and DrQ highlights the benefit of using a non-contrastive representation learning loss as an auxiliary objective for reinforement learning.

It is interesting to note that CURL, which utilizes a contrastive auxiliary loss, falls behind of all the other methods, including the ones that do not have any auxiliary loss. One reason for the observed performance drop in CURL may be the utilization of relatively small batch sizes. Since contrastive losses require negative examples, the batch size of the CURL implementation may not be enough to get good ones. On the other hand, increasing the batch size may have a detrimental effect on reinforcement learning performance after a certain point and is not always practical considering computational resources. This results in an antagonistic relationship between contrastive representation learning and reinforcement learning. Note that \cite{Agarwal2021DeepRL} used the default implementation of CuRL by \cite{Srinivas2020CURLCU}, who in turn report a batch size of 512.

By enforcing similarity in both current and future representations, SPR accentuates the performance discrepancy. The key factors contributing to SPR's effectiveness are the additional parameterization for future state prediction and the augmentations from DrQ. Consequently, BarlowRL distinguishes itself by achieving notable performance solely by leveraging information from the current state.

These results suggest that using a non-contrastive representation learning objective as an auxiliary loss improves reinforcement learning performance and is superior to contrastive objectives. As a result, our research emphasizes the importance of exploring non-contrastive objectives, as they eliminate the reliance on negative samples often obtained from randomly sampled batches. 

Lastly, we present the BarlowRL agent configuration in Appendix \ref{apd:first} and score distributions across games in Appendix \ref{apd:second}. 

\section{Future Work}
SPR ~\citep{Schwarzer2020DataEfficientRL} has demonstrated the significance of incorporating future state prediction and augmentation as a game-changing feature in representation learning for reinforcement learning. By generating multiple predictions of future states and leveraging self-supervised consistency loss, BYOL similarity, through a transition model and projection head, SPR effectively captures the temporal dynamics and dependencies crucial for reinforcement learning tasks.

Integrating non-contrastive objectives into the framework of future state prediction, similar to SPR, opens up new avenues for exploration. By formulating objective functions that directly capture the desired properties of future state predictions, such as temporal coherence or the preservation of relevant information, we can potentially improve the representation learning process. This advancement could contribute to more efficient and effective decision-making by reinforcement learning agents, as they heavily rely on accurate predictions of future states based on current state information.

Our results suggest that investigating non-contrastive objectives as auxiliary losses in other reinforcement learning methods is a potentially fruitful endeavour. Other domains, especially ones with large state spaces, are attractive targets for future work. Incorporating longer-term past dependencies, beyond the last 4 states, would be useful for partially-observable environments which can be done by utilizing recurrent layers or tranformers. Model-based RL, especially methods learning that utilize latent representations, would also benefit from representation learning. Lastly, representation learning has the potential to help with exploration as richer representations may lead to better separation of explored and unexplored states.

\section{Conclusion}

In this work, we proposed BarlowRL, a modified version of a base q-learning algorithm, that incorporates a non-contrastive objective as an auxiliary loss during batch updates. The experiments focus on evaluating BarlowRL's performance on the Atari 100k benchmark alongside other model-free reinforcement learning approaches, including one with a contrastive auxiliary loss.

BarlowRL demonstrates competitive performance across multiple metrics, outperforming DER, CURL, and DrQ in most cases. It falls slightly short compared to SPR, which utilizes future state prediction and additional augmentations. The results highlight the effectiveness of non-contrastive objectives in improving representation learning and eliminating the need for negative samples obtained from randomly sampled batches. 

By leveraging information solely from the current state, BarlowRL showcases the potential of non-contrastive objectives in reinforcement learning, in contrast to contrastive objectives which have potential to degrade performance. The findings suggest that exploring different non-contrastive objectives can lead to further advancements in sample-efficient reinforcement learning algorithms.

Overall, BarlowRL serves as a novel baseline agent and contributes to the field of sample-efficient reinforcement learning by providing a comprehensive assessment of its performance in comparison to existing model-free approaches.

\acks{This work was supported by KUIS AI Center computational resources.}

\bibliography{acml23}

\newpage
\appendix

\section{Agent Configuration}\label{apd:first}
We present the hyperparameters of the BarlowRL agent in Table~\ref{table:hyperparametersatari}. For further implementation details and for reproducing our results, feel free to check our repository here: \url{https://github.com/asparius/BarlowRL}. 

\begin{table}[h]
\caption{Hyperparameters used for Atari100K BarlowRL experiments.}
\label{table:hyperparametersatari}
\begin{center}
\begin{small}
\begin{tabular}{ll}
\toprule
\textbf{Hyperparameter} & \textbf{Value}  \\
\midrule
Random crop    & True  \\ 
Image size    & $(84,84)$  \\ 
Data Augmentation & Random Crop (Train) \\
Replay buffer size    & $100000$ \\ 
Training frames & $400000$ \\ 
Training steps & $100000$ \\
Frame skip & $4$ \\
Stacked frames    & $4$  \\ 
Action repeat    & $4$ \\
Replay period every & $1$ \\
Q network: channels & $32$, $64$ \\
Q network: filter size & $5\times 5, 5\times 5$ \\
Q network: stride & $5$, $5$ \\
Q network: hidden units & $256$ \\
Momentum (EMA for BarlowRL) $\tau$ & $0.001$  \\
Non-linearity & ReLU \\
Reward Clipping   & $[-1, 1]$  \\ 
Multi step return & $20$ \\
Minimum replay size for sampling & $1600$ \\ 
Max frames per episode & $108$K \\
Update & Distributional Double Q \\
Target Network Update Period & every $2000$ updates \\
Support-of-Q-distribution & $51$ bins \\ 
Barlow Twins $\lambda$  & $0.0051$ \\
Discount $\gamma$ & $0.99$ \\
Batch Size & $32$  \\
Optimizer    & Adam  \\ 
Optimizer: learning rate & $0.0001$ \\
Optimizer: $\beta1$ & $0.9$ \\
Optimizer: $\beta2$ & $0.999$ \\ 
Optimizer $\epsilon$ & $0.000015$ \\
Max gradient norm & $10$ \\ 
Exploration & Noisy Nets   \\
Noisy nets parameter & $0.1$ \\
Priority exponent & $0.5$ \\
Priority correction & $0.4 \rightarrow 1$ \\
Hardware & GPU \\
\bottomrule
\end{tabular}
\end{small}
\end{center}
\vskip -0.1in
\end{table}

\section{Reward Distribution Across Games}\label{apd:second}
\vspace{0.1cm}
\begin{figure*}[!h]
  \centering
  \includegraphics[width=1\textwidth]{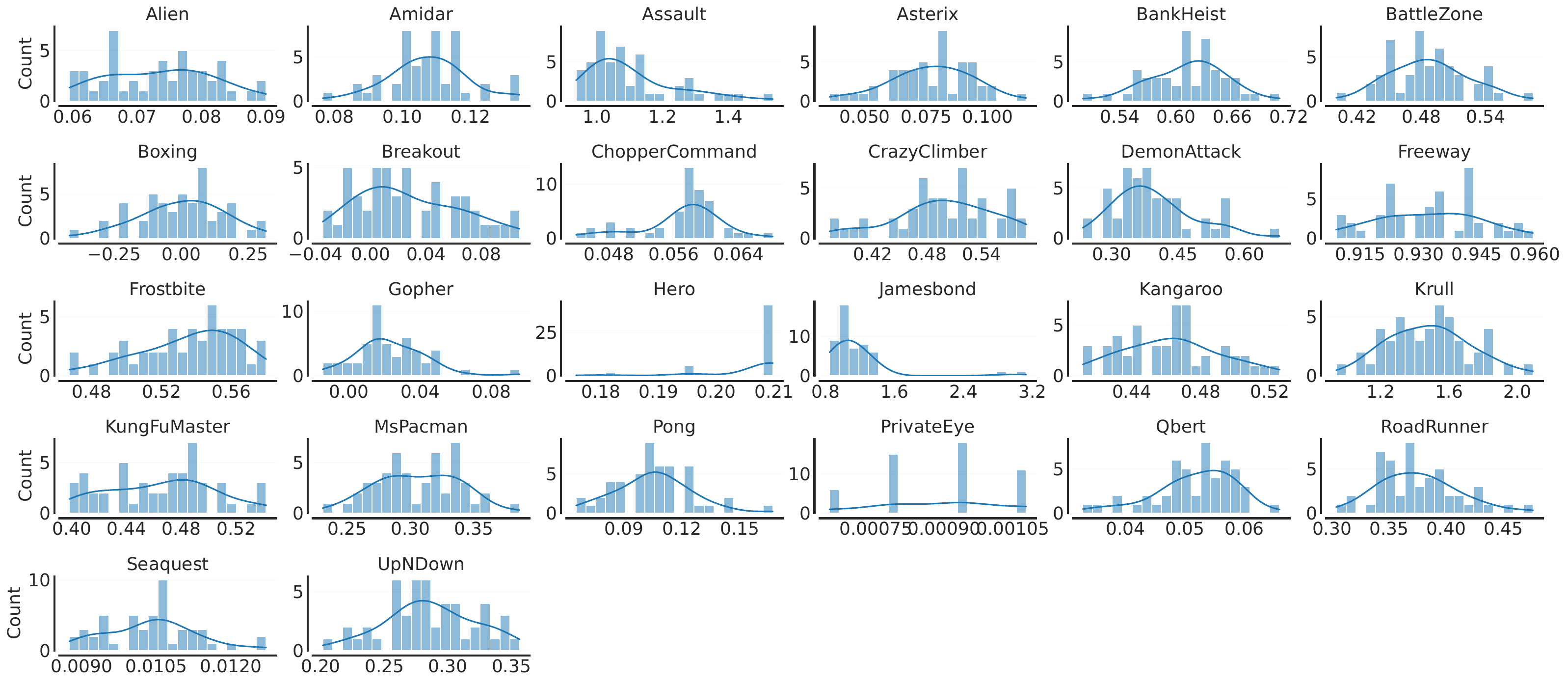}
  \caption{Distribution of rewards across 26 Atari games obtained by the BarlowRL agent.}
  \label{fig:reward}
\end{figure*}

\end{document}